\PassOptionsToPackage{table}{xcolor}
\documentclass[10pt,twocolumn,letterpaper]{article}

\usepackage[pagenumbers]{cvpr} 

\usepackage{graphicx}
\usepackage{amsmath}
\usepackage{amssymb}
\usepackage{booktabs}
\usepackage{algorithm}
\usepackage[noend]{algpseudocode}
\algnewcommand{\LineComment}[1]{\State \(\triangleright\) {\color{airforceblue} #1 } }
\makeatletter
\def\BState{\State\hskip-\ALG@thistlm}
\makeatother

\usepackage{pifont}
\newcommand{\cmark}{\ding{51}}%
\usepackage{hyperref}
\hypersetup{colorlinks,breaklinks}

\makeatletter
\@namedef{ver@everyshi.sty}{}
\makeatother
\usepackage{tikz}

\usepackage[capitalize]{cleveref}
\crefname{section}{Sec.}{Secs.}
\Crefname{section}{Section}{Sections}
\Crefname{table}{Table}{Tables}
\crefname{table}{Tab.}{Tabs.}

\usepackage{tcolorbox}

\definecolor{airforceblue}{rgb}{0.36, 0.54, 0.66}
\definecolor{darkblue}{rgb}{0.0, 0.0, 0.55}
\def\cvprPaperID{8079} 
\def\confName{CVPR}
\def\confYear{2023}

\begin{document}

\title{CPT-V: A Contrastive Approach to Post-Training Quantization of Vision Transformers}

\author{Natalia Frumkin\\
The University of Texas at Austin\\
{\tt\small nfrumkin@utexas.edu}
\and
Dibakar Gope\\
Arm Inc.\\
{\tt\small dibakar.gope@arm.com}
\and
Diana Marculescu\\
The University of Texas at Austin\\
{\tt\small dianam@utexas.edu}
}
\maketitle

\begin{abstract}
When considering post-training quantization, prior work has typically focused on developing a mixed precision scheme or learning the best way to partition a network for quantization. In our work, CPT-V, we look at a general way to improve the accuracy of networks that have already been quantized, simply by perturbing the quantization scales.
Borrowing the idea of contrastive loss from self-supervised learning, we find a robust way to jointly minimize a loss function using just $1,000$ calibration images. 
In order to determine the best performing quantization scale, CPT-V contrasts the features of quantized and full precision models in a self-supervised fashion.

Unlike traditional reconstruction-based loss functions, the use of a contrastive loss function not only rewards similarity between the quantized and full precision outputs but also helps in distinguishing the quantized output from other outputs within a given batch. 
In addition, in contrast to prior works, CPT-V proposes a block-wise evolutionary search to minimize a global contrastive loss objective, allowing for accuracy improvement of existing vision transformer (ViT) quantization schemes.
For example, CPT-V improves the top-1 accuracy of a fully quantized ViT-Base by  $10.30\%$, $0.78\%$, and $0.15\%$ for $3$-bit, $4$-bit, and $8$-bit weight quantization levels. Extensive experiments on a variety of other ViT architectures further demonstrate its robustness in extreme quantization scenarios. Our code is available at \texttt{<link>}.

\end{abstract}

\section{Introduction}
\label{sec:intro}

\begin{figure}[ht]
    \centering
    \includegraphics[width=0.8\linewidth]
                   {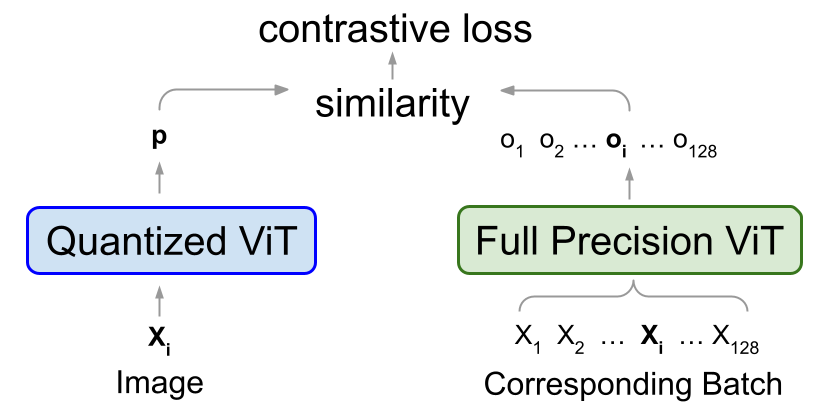}
    \caption{CPT-V uses contrastive loss to measure the similarity between a quantized prediction and the full precision predictions in the corresponding batch. Figure design is heavily inspired by MoCo~\cite{he2020momentum}.}
    \label{fig:att-grabber}
\end{figure}

Quantization is a widespread technique for efficient neural network inference: reducing data precision from $32$ bits to $\leq$ $8$ bits is an effective approach to aggressively reduce model footprint and speed up computation. Network quantization can be used for on-device inference or in cloud scenarios where some accuracy can be sacrificed for an improvement in performance. We expect these devices to have improved hardware support for quantization in the future, where $8$-bit, $4$-bit, and even $3$-bit acceleration is available for mobile devices and cloud applications. In these scenarios, full end-to-end quantization is advantageous because we can physically reduce storage and take full advantage of integer compute units. 
This paper focuses on post-training quantization of end-to-end neural networks, especially vision transfomers, where weights and activations of all layers are quantized.

\begin{figure*}
  \centering
  \begin{subfigure}{0.65\linewidth}
    \includegraphics[width=1\linewidth]
               {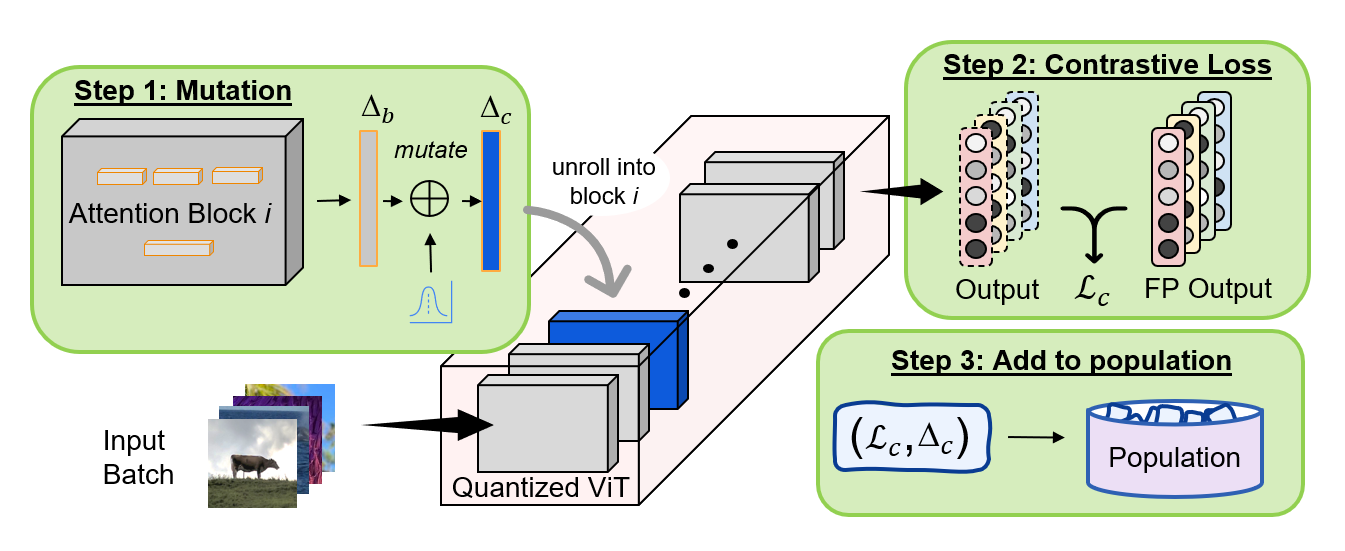}
    \caption{A single cycle of block-wise evolutionary search.}
    \label{fig:one-cycle}
  \end{subfigure}
  \hfill
   \begin{subfigure}{0.3\linewidth}
    \includegraphics[width=1\linewidth]
               {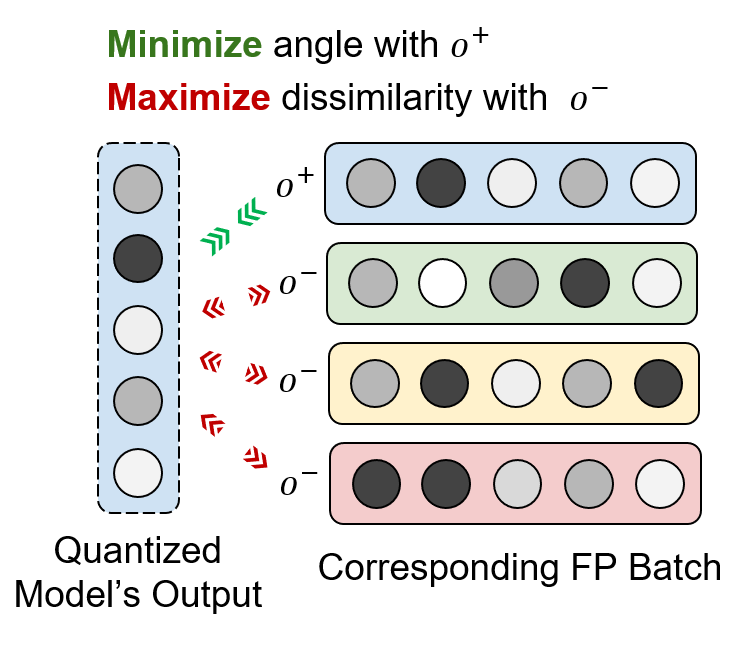}
    \caption{Visualization of Contrastive Loss (Step 2)}
    \label{fig:loss-visualization}
  \end{subfigure}
  \caption{An overview of CPT-V. On the left, we show one cycle completed on a single block. Each block has $C$ cycles of evolutionary search, and we perform $K$ passes over all blocks. On the right, we provide intuition for contrastive loss (Step 2), where we encourage similarity between the quantized and corresponding predictions while simultaneously maximizing dissimilarity between unlike predictions.}
  \label{fig:main-figure}
  
\end{figure*}

In this setting, we develop a quantization framework that aims to improve accuracy while maintaining the same quantization scheme as the baseline. \textit{Our framework, CPT-V, is the first work to use contrastive loss for neural network quantization}. In CPT-V, we show that a contrastive loss is effective in improving quantization performance, whereas traditional loss functions such as mean squared error, cosine similarity, and the KL divergence tend to overfit the calibration dataset. Using this advantage, we use block-wise evolutionary search to minimize a \textit{single objective global loss}. Ultimately, we find contrastive loss to be very effective in learning quantized representations and robust against representation collapse. The contributions of this paper are three-fold:
\begin{enumerate}
    \item \textbf{Contrastive Loss Function.} We use a contrastive loss to measure the similarity between a quantized prediction and the corresponding batch of full precision predictions. The contrastive loss minimizes the distance between the quantized and corresponding full precision predictions while simultaneously maximizing the dissimilarity with other predictions in the batch.
    \item \textbf{Block-wise Evolutionary Search.} To reduce search complexity, we employ an intuitive block-wise evolutionary search algorithm to adjust scales one block at a time. 
    \item \textbf{Global Search for Quantization Scales.} Our method finds the optimal set of quantization scales that globally minimizes a contrastive loss. By perturbing quantization scales, we find a quantized model that improves accuracy over the baseline method.
\end{enumerate}

\section{Related Work}

In this paper, we discuss techniques for \textit{post-training quantization} (PTQ). We assume no access to the initial training dataset and use a small, unlabeled calibration dataset ($1,000$ images) to help guide quantization. Typically, the calibration dataset is used for activation quantization to estimate their distributions after each layer and minimize a layer-wise loss~\cite{wu2020easyquant, hubara2021accurate,   banner2019post}. More recently, techniques involving knowledge distillation~\cite{choi2020data} and loss-aware optimization \cite{nahshan2021loss} use the calibration dataset to learn the quantization schemes of all layers jointly. However, these techniques are often victims of overfitting -- a set of $1,000$ images for a $1,000$-class dataset like ImageNet is not properly representative of the entire space of images. As a result, while using an MSE objective~\cite{jeon2022mr} or a Hessian-Aware objective~\cite{li2021brecq} may achieve reasonable results in a limited setting, they are unlikely to generalize to larger datasets and more extreme quantization where the loss landscape is highly non-smooth \cite{bai2020binarybert}. 

Alternatively, some prior works focus on layer-wise or block-wise quantization, where each layer or block is quantized separately. To learn the best quantization parameters, this work typically employs statistical analysis or a loss with respect to the full precision feature maps.
These techniques are typically generalizable to many different architectures; however, quantizing layer-by-layer may not be the best approach since one layer's quantization scheme can influence the information entropy of another layer~\cite{li2022q}.

Considering the current drawbacks of both global optimization approaches and layer-wise quantization, CPT-V attempts to learn quantization scales that are globally optimal while avoiding the overfitting problem that is common in methods that jointly optimize all quantization layers.

\begin{table*}
    \centering
    \scalebox{0.85}{
    \begin{tabular}{| c || c | c | c | c | c | c |}
    \hline
    &  Mr. BiQ & PTQ4ViT  & PSAQ-ViT & PSAQ-ViT-V2  & FQ-ViT & CPT-V  \\
    & \cite{jeon2022mr} & \cite{yuan2022ptq4vit} & \cite{choi2020data} & \cite{li2022psaq}  & \cite{lin2022fq} & [ours] \\
    \hline
    Weight Quantization & Non-Uniform & \multicolumn{4}{c|}{Uniform} & Initial range:  \\
    & Multi-Bit & \multicolumn{4}{c|}{(range: MinMax)} & MinMax \\
    \hline
    Activation Quantization & Uniform & Twin Uniform  & \multicolumn{2}{c|}{Asymmetric Uniform} &\multicolumn{2}{c|}{Log2} \\
    \hline
    Calibration Set  & 1K & 128 & \multicolumn{2}{c|}{Data-Free}  & \multicolumn{2}{c|}{1K} \\
    \hline
    Loss Type & Least-Squares & Hessian-based & - & KL & - & Contrastive\\
    \hline    
    Optimization Level& Layer-wise & Global & \multicolumn{3}{c|}{Layer-wise } & Global\\
    \hline
    Robustness to Bit-Width & 2,3,4,6  & 6,8 & 8 & 8 & 8 & 3,4,8  \\
    \hline    
    Fully Quantized & - & \cmark & \cmark & - & \cmark & \cmark  \\
    \hline 
    Optimized Quantization Scale & \cmark & - & - & - &  - & \cmark  \\
    \hline
    \end{tabular}}
    
    \caption{Comparison with Related Work. Optimization level refers to how the quantization parameters are being learned, either with a global or a layer-wise objective. We include robustness to bit-width to illustrate that each method is geared towards a different weight quantization scheme. CPT-V initializes weight quantization with MinMax, and optimizes the quantization scale to obtain the best accuracy.  }
    \label{tab:comparison}
\end{table*}

Vision Transformers (ViTs) have shown impressive results on the latest computer vision tasks \cite{touvron2021training, wu2020easyquant}. However unlike CNNs, ViT layers have extremely unbalanced distributions making quantization choice critical for achieving good predictive performance. To address this, PTQ-for-ViT~\cite{liu2021post}, learns a quantization scheme for each layer by choosing (1) the bit-width of a multi-head attention module using an attention map ranking loss and (2) the bit-width of the MLP layer using cosine similarity. This work focuses on bit-width choice, whereas other works~\cite{yuan2022ptq4vit, ding2022towards} focus on optimization of quantization scales. PTQ4ViT~\cite{yuan2022ptq4vit} employ twin-uniform quantization to improve the quantization of post-GeLU/Softmax activations and use Hessian-guided global optimization to learn the quantization scales of all linear layers. As previously stated, a Hessian-guided metric may perform well in a smooth loss landscape, such as 8-bit quantization with a well-designed dataset, but is likely to degrade in lower bit quantization schemes.
Another technique, PSAQ-ViT-V2~\cite{li2022psaq}, uses a student-teacher minmax game to minimize the KL divergence between the full precision and quantized models. See \cref{tab:comparison} for a breakdown of closely related prior works.

Our method's baseline, FQ-ViT~\cite{lin2022fq}, is a layer-wise quantization method which incorporates Log2 quantization and an integer softmax function for end-to-end $8$-bit ViT quantization. In our method, we use FQ-ViT as initialization for a contrastive loss-based global optimization scheme.

Quantization-Aware Training (QAT) has shown impressive results on vision transformers~\cite{xu2022tervit, li2022q, li2022auto, li2022vit}, yet these methods consider training with the entire dataset rather than an unlabeled calibration dataset. We provide a comparison to PTQ methods using QAT-based methods as an ``oracle." 

Currently, there is limited work that looks at combining quantization and self-supervised learning. These works ~\cite{cao2022synergistic, fu2022contrastive} use quantization-aware training in conjunction with a self-supervised learning scheme. They claim that quantization allows for regularization during training and is complementary to using augmentation for self-supervised learning. Both contributions ~\cite{cao2022synergistic, fu2022contrastive} combine a quantization-based loss and a contrastive loss into a joint optimization scheme, however, they do not use it in the same way we do in this paper. They apply a contrastive loss with respect to only full precision predictions, whereas we apply the contrastive loss as a reconstruction loss for the quantized predictions (see \cref{fig:loss-visualization}).

\section{ The CPT-V Framework}

CPT-V is a post-training technique for end-to-end quantization. We quantize both the weights and activations for all layers, as done in \cite{lin2022fq}. In particular, we quantize all activations, including those around the Softmax and LayerNorm layers. To improve the accuracy of an existing quantized model, we use a block-wise evolutionary search algorithm to search for the set of scales which are globally optimal.

\subsection{Uniform, End-to-End Quantization}
We consider uniform quantization, where the quantization range is between $\alpha$ and $\beta$, initialized using MinMax~\cite{jacob2018quantization} for weights and Log2~\cite{cai2018deep} for the activations. Uniform quantization is formally defined as:

\begin{equation}
  Q(\textbf{x}, \delta, \alpha, \beta) = clip(round( \frac{\textbf{x}}{\delta}), \alpha, \beta)
  \label{eq:quant}
\end{equation}

where $\textbf{x}$ is a full-precision vector and  $\delta$ is the quantization scale.  In our framework, we learn these initial quantization parameters using a fast, layer-wise framework such as FQ-ViT~\cite{lin2022fq}, and then use evolutionary search to adjust the scales of each attention block.

\subsection{Global Search for Quantization Scales}

When considering \textit{global optimization} of all quantization scales, the dimensionality of the search space becomes extremely high. Each weight matrix and attention map may be assigned many quantization parameters (\eg, token-wise or group-wise quantization), and the number of scales may grow to over $10,000$ per model. In this case, the search space of all quantization scales is typically partitioned in a block-wise or layer-wise fashion. Similar to \cite{li2021brecq}, we partition our search space in a block-wise manner, but evaluate each block using a global fitness metric based on a contrastive loss (see \cref{sec:fitness} for more details).
\begin{tcolorbox}
[width=\linewidth, sharp corners=all, colback=white!95!black]
\textit{\textbf{Q: Why not use gradient descent?}}

The quantization function is non-differentiable. Normal quantization-aware training typically circumvents this by training the weights as opposed to the quantization scales and using a straight-through estimator. In order to adjust the scales directly, we choose to use evolutionary search since it converges fast enough in this setup.
\end{tcolorbox}

\begin{algorithm}
\caption{Block-wise Evolutionary Search}\label{joint-evol}
\hspace*{\algorithmicindent} \textbf{Input:} Calibration Dataset $D_{C}$ \\
\hspace*{\algorithmicindent} \hspace{3mm} Quantized Model $M_{Q}$, Full Precision Model $M_F$ \\
\hspace*{\algorithmicindent} \hspace{3mm} Number of passes $K$, cycles $C$ \\
\hspace*{\algorithmicindent} \hspace{3mm} Population size $P$, Sample size $S$
\begin{algorithmic}[1]
\LineComment{traverse all attention blocks}
\For{ pass in 0 : $K$}
\For{ $b$ in \texttt{AttentionBlocks}($M_{Q}$)}
    \LineComment{begin search for attention block $b$}
    \State pop $\gets$ [\hspace{1mm} ]
    \While {$|$ pop $|$ $<$ $P$ }
    \Comment{initialize population}
        \State fitness $\gets$ \texttt{Fitness}($b$, $M_Q$, $M_F$, $D_{C}$)
        \State pop\texttt{.insert(($\boldsymbol\Delta_{\texttt{b}}$, fitness))}
    \EndWhile
    \For{ cycle in 0 : $C$}
        \LineComment{choose parent scales from sampled set}
        \State samples $\gets$ [\hspace{1mm} ]
        \While {$|$ samples $|$ $<$ $S$ }
            \State samples $\gets$ \texttt{RandomElement}(pop)
        \EndWhile
        \State parent $\gets$ \texttt{BestFitness}(samples)
        \State
        \LineComment{evaluate child and add to population}
        \State $\boldsymbol\Delta_{\texttt{child}}$ $\gets$ \texttt{Mutate}($\boldsymbol\Delta_{\texttt{parent}}$)
        \State fitness $\gets$ \texttt{Fitness}($D_C$, $M_Q$, $M_F$ )
        \State pop\texttt{.insert}( ($\boldsymbol\Delta_{\texttt{child}}$, fitness ) )
        \LineComment{remove candidate with lowest fitness}
        \State pop.\texttt{DeleteDead}()
    \EndFor
    \State $\boldsymbol\Delta_{\texttt{b}}$ $\gets$ \texttt{BestFitness}(pop)
\EndFor
\EndFor
\State \Return $M_{Q}$
\Comment model with updated scales
\end{algorithmic}
\end{algorithm}

In order to find the optimal scales for all layers, we employ block-wise evolutionary search, as shown in \cref{joint-evol}. We represent all scales in an attention block as a stacked 1-D vector $\boldsymbol\Delta_{b}$, and perturb this vector during search to find the best scales with respect to the fitness function (evaluation metric). Once we have completed our search for one block, we go on to the next block and repeat this process. 

For each attention block, we apply a small evolutionary search procedure to learn the best quantization scales within a neighborhood of the current scales. This sub-problem consists of $C$ cycles of evolutionary search, where the scales are mutated once per cycle in the following manner:

\begin{equation}
  \boldsymbol\Delta_{\texttt{child}} \sim \mathcal{U}(\boldsymbol\Delta_{\texttt{parent}}, -\gamma, \gamma)
  \label{eq:mutate}
\end{equation}

Our mutation function, \cref{eq:mutate}, generates a new child scale, $\boldsymbol\Delta_{\texttt{child}}$, from a uniform distribution parameterized by $\gamma$. This child scale is assigned to the current block $b$ and evaluated using our fitness function. 

We apply this small evolutionary search sub-problem to a given block $b$ and traverse through all blocks to jointly learn all quantization parameters. $P$ passes are performed over all attention blocks (\ie, $K$ passes means search is applied to each block $K$ times). Through trial and error, we find that $P >> C$ is most effective, demonstrating that we want many smaller evolutionary search problems rather than fewer large search problems.

\subsection{Contrastive Loss}
\label{sec:fitness}

Contrastive loss is commonly used in self-supervised learning to prevent representation collapse \cite{grill2020bootstrap} and to encourage discrimination between a target representation and a set of negative examples. We find a contrastive loss to be very effective in preventing the collapse of a \textit{quantized network's} representation in the same way that it is used to develop richer representations in a self-supervised setting. Inspired by \cite{chen2021empirical}, we use the infoNCE~\cite{oord2018representation} loss:

\begin{equation}
  \mathcal{L}_{c} = - \log \frac{\exp(p \cdot o^{+} / \tau) }{\exp(p \cdot o^{+} / \tau) + \sum_{o^{-}} \exp(p \cdot o^{-} / \tau)}
  \label{eq:infoNCE}
\end{equation}

where $p$ is the prediction of the quantized model, $o^{+}$ is the corresponding prediction of the full precision model, and $o^{-}$ is a prediction of another image in the same batch. We use infoNCE in a novel way: to improve the reconstruction of $o^{+}$ while enforcing that the quantized model prediction is dissimilar from other images in a batch.

\begin{algorithm}
\caption{Fitness Function}\label{fitness}
\hspace*{\algorithmicindent} \textbf{Input:} calibration dataset $D_C$ \\
\hspace*{\algorithmicindent} \hspace{3mm} quantized model $M_Q$, full precision model $M_F$
\begin{algorithmic}[1]
\For{ batch in $D_{C}$}
\State p = $M_Q$(batch)
\State o = $M_F$(batch)
\State score += \texttt{ContrastiveLoss}(p, o)
\Comment{\cref{eq:infoNCE}}
\EndFor
\State \Return score / \texttt{size}($D_C$)
\end{algorithmic}
\end{algorithm}

When evaluating a block's fitness, we take the average contrastive loss with respect to the calibration dataset as shown in \cref{fitness}. We emphasize that fitness is a \textit{global evaluation metric} to evaluate how well $\boldsymbol\Delta_{b}$ affects the final prediction, and is not evaluated on any intermediate feature map individually.

\section{Results}
In the following section, we present results on a variety of vision transformers and show the consistency of our method under standard $8$-bit quantization and in extreme quantization schemes ($3$-bit and $4$-bit weights). We present results for end-to-end quantization, where all weights and activations are quantized. For all experiments, we consider 8-bit activation quantization and vary weight quantization.
\subsection{Setup}
We conduct experiments using ImageNet (ILSVRC2012) and evaluate our post-training techniques on a variety of vision transformer model families. The calibration dataset is $1,000$ randomly sampled images from the ImageNet training set (without labels). We note that this is a different setup from quantization-aware training, where the entire training dataset is used. The initial quantized model ($M_Q$ in \cref{joint-evol}) is generated using FQ-ViT, and our method does not adjust any quantization settings. It only perturbs the quantization scales. FQ-ViT is an end-to-end quantization framework that uses MinMax~\cite{jacob2018quantization} for weight quantization and Log2~\cite{cai2018deep} for activation quantization. We refer to FQ-ViT and our code for all other quantization settings.

For block-wise evolutionary search, we specify all search settings in \cref{tab:settings} below.

\begin{table}[h]
    \centering
    \begin{tabular}{ c c|c }
    \toprule
         passes & $K$ & 10 \\
         population size & $P$ & 15 \\
         cycles & $C$ & 3 \\
         samples & $S$ & 10 \\
         mutation range & $\gamma$ & $10^{-3}/10^{-4}$ \\
    \bottomrule
    \end{tabular}
    \caption{Block-wise evolutionary search settings. The mutation range is $10^{-3}$ for 8W8A, and $10^{-4}$ for 4W8A and 3W8A.}
    \label{tab:settings}
\end{table}

\subsection{8-bit Quantization}

We compare our standard 8-bit quantization with state-of-the-art methods in \cref{tab:8w8a}.  We improve over existing \textit{end-to-end} quantization techniques by at least $0.1\%$, $1.2\%$, and $0.15\%$ for DeiT-Small, DeiT-Base, and ViT-Base, respectively. 

\begin{table}[h]
  \centering
  \rowcolors{2}{gray!25}{white}
  \begin{tabular}{| c | c  c  c  c |}
    \hline
    \multicolumn{5}{|c|}{8-bit weights, 8-bit activations (8W8A)} \\
    \hline
    Method & DeiT-T & DeiT-S & DeiT-B & ViT-B \\
    \hline
    PSAQ-ViT & 71.56 & 76.92 & 79.10 & 37.36 \\
    PTQ4ViT  & - & 79.47 & 81.48 & 84.25 \\
    FQ-ViT  & 71.61 & 79.17 & 81.20 & 83.31 \\
    PSAQ-ViT-V2\textsuperscript{\textdagger} & \textbf{72.17} & 79.56 & 81.52 & -  \\
    CPT-V (ours) & 71.63 & \textbf{79.57} & \textbf{82.67} & \textbf{84.40} \\
    \hline
  \end{tabular}
  
  \footnotesize{\textsuperscript{\textdagger} Does not quantize Softmax/GELU layers}
  \caption{Top-1 Accuracy on ImageNet using 8W8A quantization. -T, -S, \& -B refer to Tiny, Small, and Base models respectively.}
  \label{tab:8w8a}
\end{table}

We also compare with PSAQ-ViT-V2~\cite{li2022psaq} and find that it outperforms our method by $0.5\%$ for DeiT-Tiny. However, PSAQ-ViT-V2 is not a full end-to-end quantization method since it \textit{does not quantize the activations} following the Softmax and GELU layers. These activations are typically very sensitive to quantization and are often maintained at full precision for this purpose. In our work, we consider end-to-end quantization, so we are forced to quantize the post-Softmax/GELU activations. We leave it to future work to apply our technique on top of PSAQ-ViT-V2, but we expect similar improvements to what we achieved with FQ-ViT.

\subsection{4-bit Quantization}

Moving from $8$-bit to $4$-bit weight quantization, we see an accuracy degradation of about $2-5\%$ across all models. In \cref{tab:4w8a}, we find that our method performs similarly to what is shown for 8-bit quantization. In particular, we still see improvement for DeiT-Small, DeiT-Base, and ViT-Base, but now the top-1 accuracy improvement is $0.9\%$, $0.65\%$, and $0.8\%$, respectively.

\begin{table}[h]
  \centering
  \rowcolors{2}{gray!25}{white}
  \begin{tabular}{| c | c  c  c  c |}
    \hline
    \multicolumn{5}{|c|}{4-bit weights, 8-bit activations (4W8A)} \\
    \hline
    Method &  DeiT-T & DeiT-S & DeiT-B & ViT-B \\
    \hline
    PSAQ-ViT & 65.57 & 73.23 & 77.05 & 25.34 \\
    PTQ4ViT  & - & - & 64.39 & - \\
    FQ-ViT  & 66.91 & 76.93 & 79.99 & 78.73 \\
    PSAQ-ViT-V2\textsuperscript{\textdagger}  & \textbf{68.61} & 76.36 & 79.49 & -  \\
    CPT-V (ours) & 67.29 & 77.06 & 80.15 & \textbf{79.50} \\
    \hline
  \end{tabular}
  \footnotesize{\textsuperscript{\textdagger} Does not quantize Softmax/GELU layers}
  \caption{Top-1 Accuracy on ImageNet using 4W8A quantization.}
  \label{tab:4w8a}
\end{table}

\subsection{3-bit Quantization}

We report 3-bit quantization results in \cref{tab:3w8a} to show that CPT-V extends to more extreme quantization scenarios. In particular, CPT-V improves accuracy over FQ-ViT by $10.3\%$ for ViT-Base and $3.7\%$ for DeiT-Tiny. We compare with Mr.BiQ~\cite{jeon2022mr}, a PTQ technique that uses non-uniform quantization and does not quantize the Softmax/GELU layers. 

\begin{table}[h]
  \centering
  \rowcolors{2}{gray!25}{white}
  \begin{tabular}{| c |  c  c  c  c |}
    \hline
    \multicolumn{5}{|c|}{3-bit weights, 8-bit activations (3W8A)} \\
    \hline
    Method &  DeiT-T & DeiT-S & DeiT-B & ViT-B \\
    \hline
    FQ-ViT &  35.79 & 60.58 & 72.11 & 55.33 \\
    CPT-V (ours)  & \textbf{39.45} & 61.16 & 72.41 & 65.63 \\
    Mr. BiQ\textsuperscript{\textdagger}  & - & \textbf{78.09} & \textbf{81.10} & \textbf{77.41}  \\
    \hline
  \end{tabular}

  \footnotesize{\textsuperscript{\textdagger}  Non-uniform quantization, Does not quantize Softmax/GELU layers.}
  \caption{Top-1 Accuracy on ImageNet using 3W8A quantization.}
  \label{tab:3w8a}
\end{table}

By reducing the precision from $32$ to $3$ bits, we achieve a $>$10X reduction in memory footprint while still maintaining a reasonable accuracy for DeiT-Base. While DeiT-Tiny and DeiT-Small achieve low accuracy in this regime, their improvement may be drastically improved with other techniques such as OMSE~\cite{choukroun2019low}, bias correction~\cite{banner2019post}, and group-wise quantization~\cite{shen2020q}.

\subsection{Extending to Swin \& LeViT Models}

We run our experiments on additional model families to ensure that our method is applicable to different types of transformer blocks.
Swin transformers~\cite{liu2021swin} have the same macro-architecture as DeiT and ViT models, with the exception that the swin transformer block is a windowed attention. We see results for 4-bit Swin transformers in \cref{tab:swin}.

\begin{table}[h]
  \centering
  \rowcolors{2}{gray!25}{white}
  \begin{tabular}{| c |  c  c  c |}
    \hline
    Method &  Swin-T & Swin-S & Swin-B \\
    \hline
    PSAQ-ViT & 71.79 & 75.14 & - \\
    PSAQ-ViT-V2\textsuperscript{\textdagger} & 76.28 & 78.86 & - \\
    FQ-ViT & \textbf{80.73} & 82.13 & 82.73 \\
    CPT-V (ours) & 80.43 & \textbf{82.63} & \textbf{83.07} \\
    \hline
  \end{tabular}
  
  \footnotesize{\textsuperscript{\textdagger} Does not quantize Softmax/GELU layers} \\
  \caption{Top-1 Accuracy for Swin Models using 4W8A.}
  \label{tab:swin}
\end{table}

Prior quantization techniques do not consider LeViT models, a family of vision transformers that improve the inference speed of existing transformers. Using a longer convolutional backbone, they can achieve similar results to classic vision transformers while also reducing the complexity of the transformer blocks in favor of ResNet-like stages. We include the LeViT family in our experiments to illustrate how our method can be extended beyond the standard block size. We can see 4W8A results for the LeViT model family in \cref{tab:levit}. Across the board, we see a significant improvement in LeViT quantization as compared to FQ-ViT, our baseline. 
\begin{table}[h]
    \centering
    \rowcolors{2}{gray!25}{white}
    \begin{tabular}{|c|c c |}
        \hline
        Model & FQ-ViT & CPT-V (ours) \\
        
        \hline
        LeViT-128S & 14.90 & \textbf{29.20} \\
        LeViT-192 & 17.00 & \textbf{30.37}\\
        LeViT-256 & 61.33 & \textbf{64.57}\\
        LeViT-384 & 64.60 & \textbf{69.50}\\
        \hline
    \end{tabular}
    \caption{Top-1 Accuracy for LeViT models using 4W8A.}
    \label{tab:levit}
\end{table}

In fairness, we have not applied any other techniques to boost LeViT's accuracy (doing so may inflate our method's improvement), so we leave it to future work to incorporate other quantization techniques on top of our framework.

\section{Analysis}
In the section below, we provide a discussion on runtime, various ablations, and explainability using attention maps. Our goal is to motivate the need for a fast, yet reliable technique to improve end-to-end quantization of  ViTs.

\subsection{Loss Function Choice}

\begin{figure}[h]
  \centering
\includegraphics[width=1\linewidth]
               {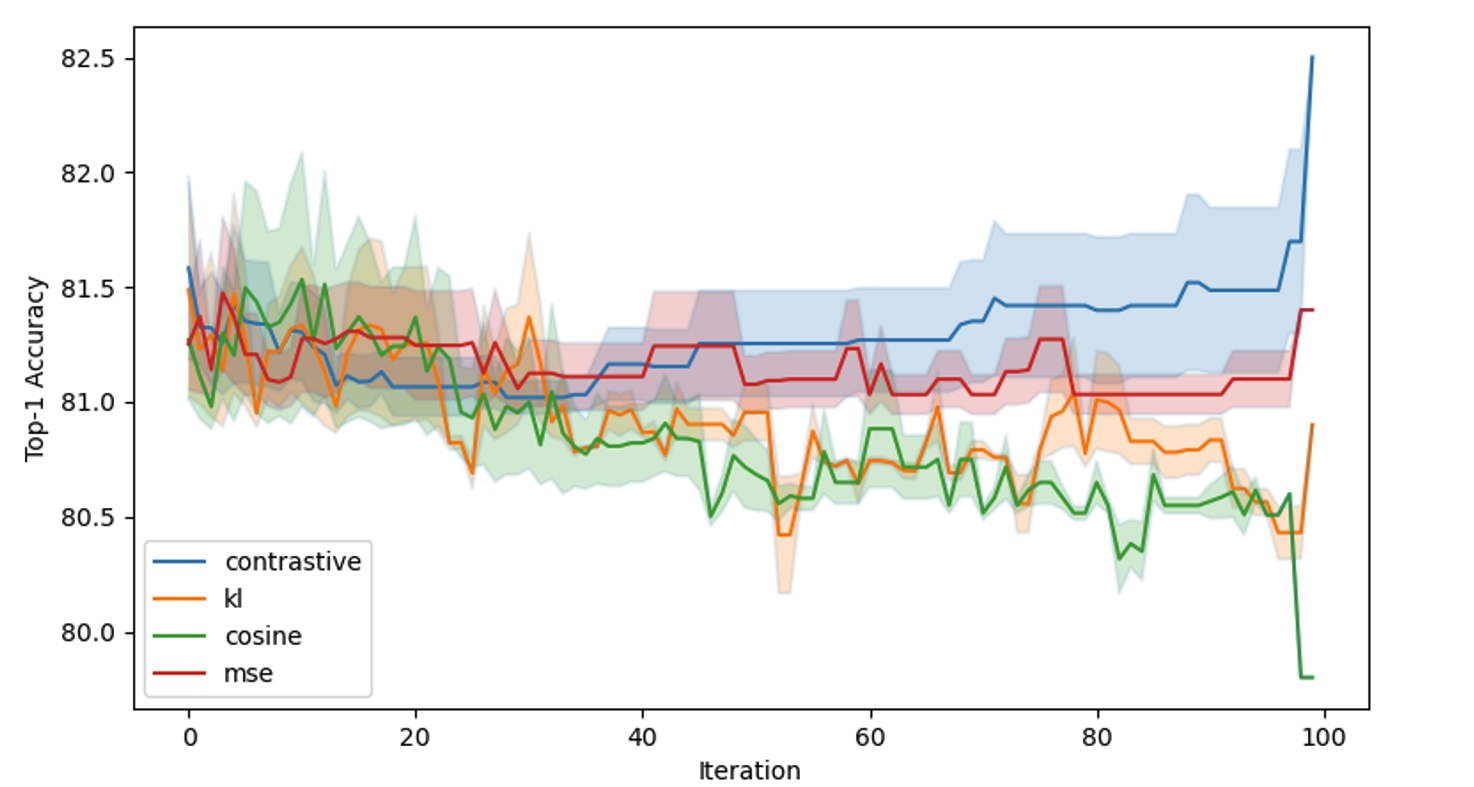}
\caption{Comparing performance of CPT-V for the four loss functions. The contrastive loss prevents overfitting to the calibration dataset, whereas all other loss functions cannot improve accuracy beyond the initialized quantization scheme.}
\label{fig:loss_types}
\end{figure}

We compare contrastive loss with other common  loss functions in \cref{fig:loss_types}. We find mean-squared error (MSE) to be equally (if not more) effective in the initial iterations of CPT-V. However, as the number of passes grows, MSE does not perform as well as the contrastive loss. Both cosine similarity and the Kullback–Leibler divergence (KL) fail to improve performance as the number of iterations increases. We postulate that the poor performance of these traditional loss functions is due to overfitting to the calibration dataset. On the other hand, contrastive loss is naturally regularized by the negative samples in the batch, allowing for the contrastive loss to preserve the quantization parameters that help discriminate between classes.

\subsection{Layer-wise Weight Distributions}

\begin{figure}[h]
    \centering
    \includegraphics[width=1\linewidth] {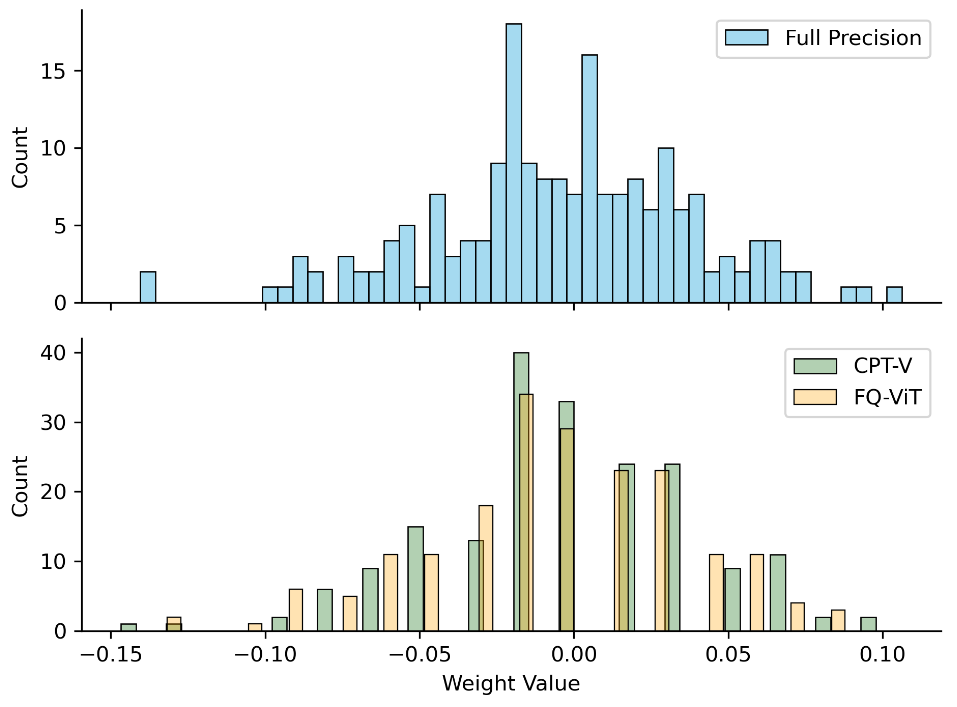}
    \caption{ The weight distribution of the projection layer for attention block \#1 of ViT-Base. The top plot shows the full precision weight distribution and the bottom plot shows the 8-bit weights using both FQ-ViT and CPT-V.}
    \label{fig:mod1_hist}
\end{figure}

In \cref{fig:mod1_hist}, we compare the weight distributions of the full precision, FQ-ViT, and CPT-V quantization schemes. CPT-V's quantized weight distribution is similar to FQ-ViT, yet  CPT-V has a $0.8\%$ improvement over FQ-ViT (note: $\%$ improvement is for this run only, \cref{tab:8w8a} is averaged over three seeds).

\begin{figure*}
  \centering
    \includegraphics[width=1\linewidth]
           {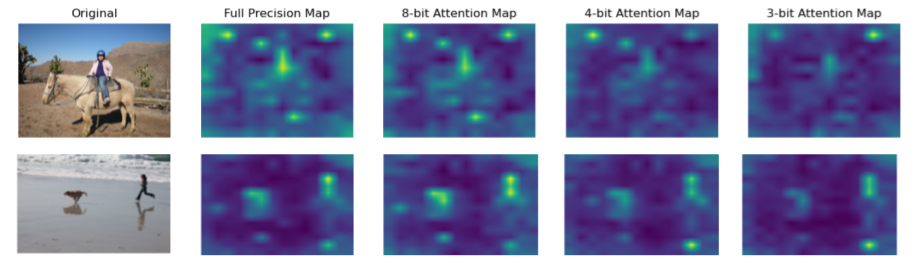}

  \caption{Attention maps for different quantization levels. CPT-V's quantized models preserve the spacial locality of the full precision feature map. As the quantization level becomes more extreme, the attention map becomes subject to decreased resolution.}
  \label{fig:att_maps}
  
\end{figure*}

Interestingly, this graph illustrates that a small perturbation in the scales can significantly improve performance. This is consistent with the conclusion in AdaRound~\cite{nagel2020up}, where they show that a perturbation in weights may yield no difference in quantized outputs. AdaRound claims that round-to-nearest is not the most effective rounding scheme since it may round away weights which are important for deciphering classes. By adjusting quantization scales, we can overcome the aforementioned rounding issue by encouraging dissimilarity between two  unlike predictions.

~\cref{fig:mod1_hist} is just one block, and we refer to the supplementary materials for a more complete discussion of how these perturbations change across different layers.

\subsection{Impact on Attention Maps}
We find that CPT-V preserves the spatial integrity of the full precision feature maps even as quantization forces discretization of the attention mechanism. In \cref{fig:att_maps}, as quantization becomes more severe from 8-bit to 3-bit, the resolution of the feature map degrades, as is expected when only a finite number of values can be expressed in the quantized scheme. This attention map visualization is averaged over all blocks, and serves as qualitative inspection of how the network's attention mechanism is performing. All in all, \cref{fig:att_maps} provides confidence that CPT-V's quantized attention maps learn reasonable representations of the original full precision network.

\subsection{Ablation: Passes vs. Cycles}
\begin{figure}[h]
    \centering
    \includegraphics[width=0.9\linewidth]
           {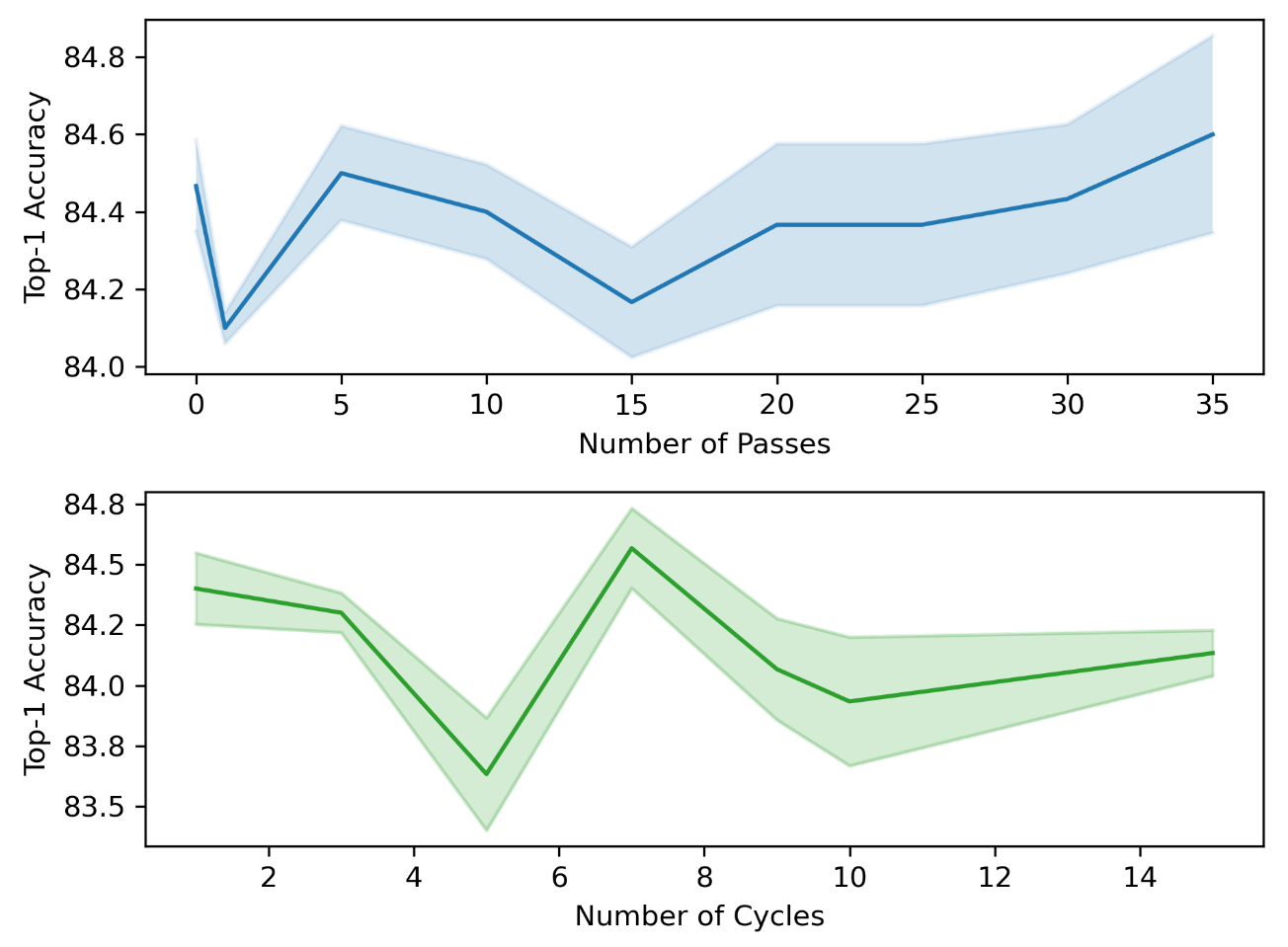}
    \caption{Ablation on number of cycles and passes}
    \label{fig:pass_cycle_ablation}
\end{figure}

In \cref{fig:pass_cycle_ablation}, we ablate the number of passes, $K$, from $1$ to $35$. As we can see, a majority of the accuracy improvement occurs in the first $10$ passes, so we choose $K = 10$ for all experiments above. This allows for our method to run in less than one hour. However, we note that an additional accuracy boost may be enjoyed with more passes. 

We also ablate the number of cycles, $C$, to determine how many mutations should occur per block (one cycle is shown in \cref{fig:one-cycle}). We use $C = 3$ even though we see $C = 7$ is optimal in our ablation study. In practice, we find that the choice of $C$ is random seed and model dependent. We find that for some runs, the best choice is simply $1$ cycle, but in others it is $3$, $5$ or $7$. Ultimately, we choose $C = 3$ for consistency across experiments. 


\subsection{Runtime}

We run our method on an Nvidia A100-PCIE-40GB Tensor Core GPU and find that all experiments take less than one hour to run. The average runtime is shown in \cref{tab:runtime}. We use Pytorch 1.9.1, built with the CUDA 11.1 toolkit.

\begin{table}[h]
  \centering
  \begin{tabular}{c  c  c  c  c }
    \toprule
    & DeiT-T & DeiT-S & DeiT-B & ViT-B \\
    Runtime (mins) & 41.5 & 46.3 & 41.6 & 43.2 \\

    \bottomrule
  \end{tabular}
  \caption{CPT-V Runtime (in minutes) on one Nvidia A100 GPU}
  \label{tab:runtime}
\end{table}

In \cref{fig:runtime}, we make a runtime comparison with other ViT quantization methods and demonstrate that our method achieves superior accuracy on ViT-Base network while being faster than QAT methods. \cref{fig:runtime} only captures the Top-1 Accuracy of ViT-Base (or, alternatively, DeiT-base if ViT-Base is unavailable). We note that there would be a different plot if we used DeiT-Tiny, since PSAQ-ViT-V2 outperforms CPT-V in this scenario. We refer to the supplementary material for a wider discussion on how this plot changes with different models. 

\subsection{Comparison with QAT Methods}
\label{sec:qat}

\begin{figure}[h]
  \centering
    \includegraphics[width=1\linewidth]
           {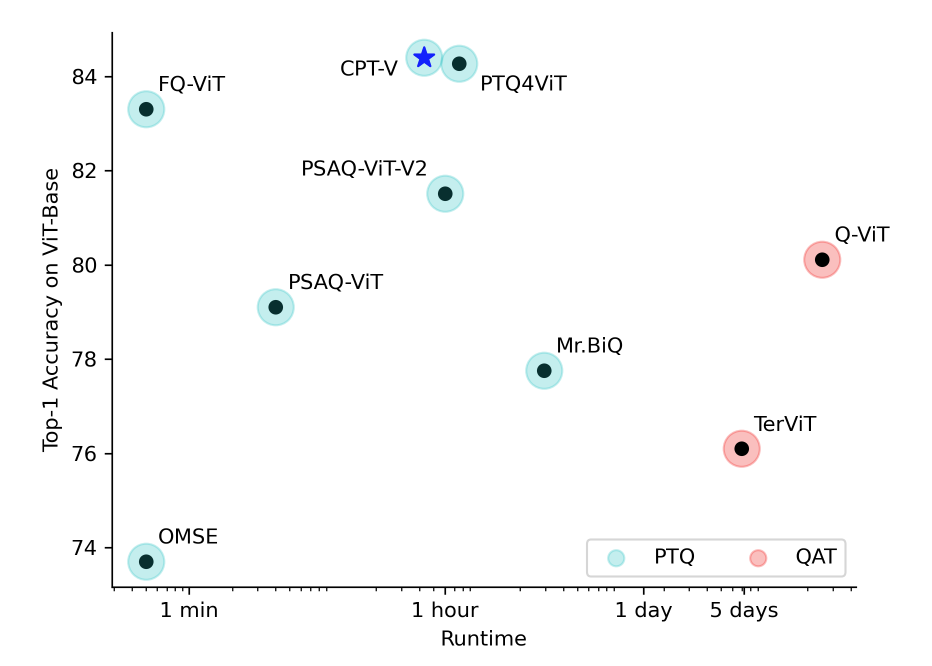}

  \caption{Runtime of various quantization methods for ViTs. We consider PTQ and QAT methods in our analysis.}
  \label{fig:runtime}
  
\end{figure}

CPT-V improves over existing post-training quantization (PTQ) and quantization-aware training (QAT) techniques for vision transformers. In \cref{fig:runtime}, CPT-V is pareto optimal in terms of both top-1 accuracy and runtime for ViT-Base for all ViT-specific quantization methods. Q-ViT~\cite{li2022q}, TerViT~\cite{xu2022tervit}, and Mr. BiQ~\cite{jeon2022mr} use only 3/4-bit weights which accounts for why they do not achieve similar top-1 accuracy as the other PTQ schemes which use 8-bit weight quantization. 



Most of these methods are run on a single Nvidia A100 GPU, but some were not open-sourced at the time of this publication. In particular,  TerViT~\cite{xu2022tervit} is estimated using LSQ's~\cite{esser2019learned} open-source ResNet-50 QAT scheme, and Mr.BiQ's runtime is reported on a Nvidia V100. There was no reported runtime for PSAQ-ViT-V2, so we estimate it to be $60$ minutes based on PSAQ-ViT and our best guess at the runtime of the additional steps.

\subsection{On Variation across Random Seeds}
\begin{figure}[h]
    \centering
    \includegraphics[width=0.9\linewidth]
                   {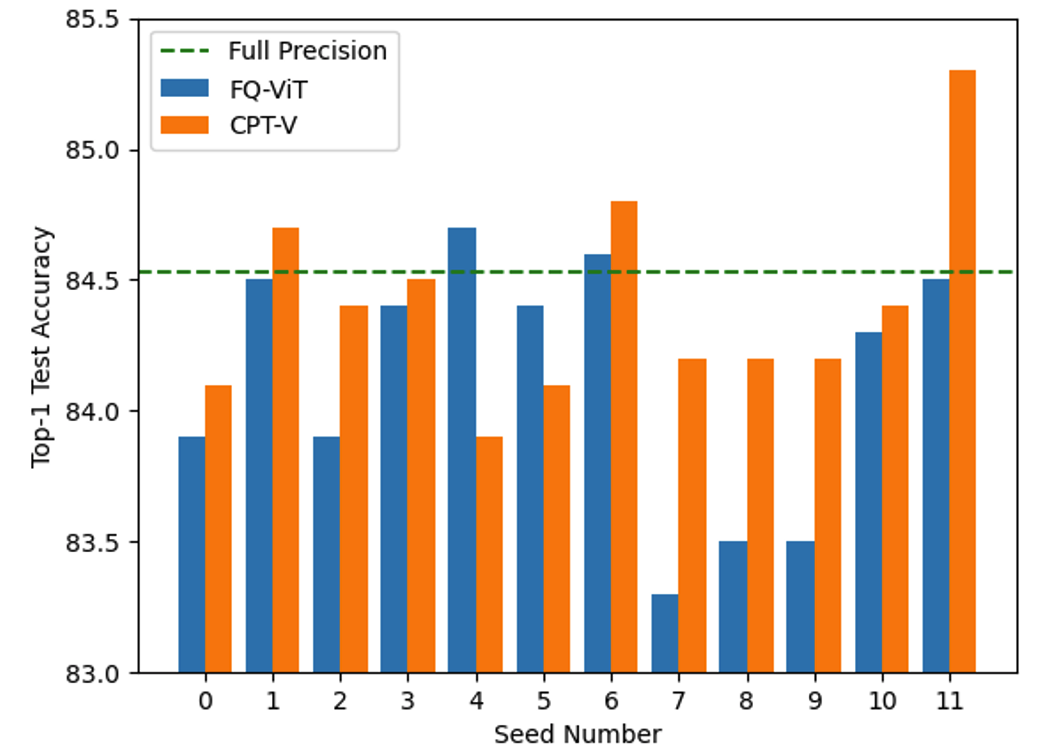}
    \caption{Comparing performance across 12 random seeds for 8W8A ViT-Base. 10/12 runs improve over the initial FQ-ViT quantization.}
    \label{fig:seeds}
\end{figure}

In \cref{fig:seeds}, we show the performance of CPT-V compared to the baseline method, FQ-ViT. Across twelve random seeds, ten runs improve performance over FQ-ViT, and three result in top-1 accuracy that is superior to the full precision model. This does not typically happen for post-training methods since no training is involved. However, we find that CPT-V can traverse through the optimization space and happen upon quantization parameters that can improve performance drastically. Overall, CPT-V has an 83\% (10/12) chance of improving on FQ-ViT for ViT-Base.
  
As with many advanced post-training methods, the random seed is a significant factor in determining the final quantization accuracy. The random seed dictates which images are chosen for the calibration dataset, as well as the way the dataset is generated. We postulate that the performance of CPT-V can be improved by choosing the proper set of images for the calibration dataset (\ie, those that are representative of the test set). For fair comparison, our experiments allow for all types of calibration sets to be generated for quantization, and we attribute the poor accuracy in seeds 4 and 5 to the poor choice of calibration set.


\subsection{Generalizability}
Our method requires a pre-quantized model, $M_Q$, and adjusts the models scales to improve accuracy. We only require that the model can be abstracted into blocks, with each block having a set of quantization scales. This makes our method readily applicable to other types of models such as CNNs, LSTMs, Graph Neural Networks, and many others. We focus exclusively on ViTs in this work, since we find them to be very successful for image classification tasks, and leave to future work the extension of this framework to other block-wise separable networks.

\section{Conclusion}
We propose CPT-V, a quantization method that uses a contrastive loss to jointly optimize the quantization scales of all attention blocks. Using block-wise evolutionary search, CPT-V improves the accuracy of an existing quantized model by minimizing a global fitness function. Using our method, we improve top-1 accuracy for 8-bit, 4-bit, and 3-bit quantization levels when compared to existing state-of-the-art ViT quantization techniques. Finally, CPT-V benefits from generality -- we can apply it to many types of blocks with different quantization schemes. As long as there are quantization scales, our method is able to optimize using block-wise evolutionary search.



\newpage

{\small
\bibliographystyle{ieee_fullname}
\bibliography{references}
}

\end{document}